\documentclass[10pt,twocolumn,letterpaper]{article}
\pdfoutput=1

\usepackage[final]{pdfpages}
\usepackage{cvpr}
\usepackage{times}
\usepackage{epsfig}
\usepackage{graphicx}
\usepackage{amsmath}
\usepackage{amssymb}
\usepackage{multirow}
\usepackage{adjustbox}
\usepackage[table,x11names]{xcolor}
\usepackage{caption}
\usepackage{subcaption}
\usepackage{dirtytalk}
\usepackage[symbol]{footmisc}
\usepackage{soul}
\setstcolor{magenta}
\usepackage[colorlinks = true,
            linkcolor = blue,
            urlcolor  = blue,
            citecolor = blue,
            anchorcolor = blue]{hyperref}

\newcommand{\camready}[1]{\textcolor{black}{#1}}
\usepackage{enumitem}

\setitemize[0]{leftmargin=10pt}

\newcommand{\MYhref}[3][blue]{\href{#2}{\color{#1}{#3}}}%

\cvprfinalcopy 

\def\cvprPaperID{8463} 

\ifcvprfinal\fi

\begin{document}

\title{
SharinGAN: Combining Synthetic and Real Data for Unsupervised Geometry Estimation
}

\author{Koutilya PNVR\\
{\tt\small koutilya@terpmail.umd.edu}
\and
Hao Zhou\footnotemark\hspace{1cm}\\
{\tt\small hzhou@cs.umd.edu}
\and
David Jacobs\\
{\tt\small djacobs@umiacs.umd.edu}
\and
University of Maryland, College Park, MD, USA.
}

\maketitle
\footnotetext{$^*$Hao Zhou is currently at Amazon AWS. }

\begin{abstract}
   We propose a novel method for combining synthetic and real images when training networks to determine geometric information from a single image.  We suggest a method for mapping both image types into a single, shared domain.  This is connected to a primary network for end-to-end training.  Ideally, this results in images from two domains that present shared information to the primary network.  Our experiments demonstrate significant improvements over the state-of-the-art in two important domains, surface normal estimation of human faces and monocular depth estimation for outdoor scenes, both in an unsupervised setting. 
   
\end{abstract}

\section{Introduction}

Understanding geometry from images
is a fundamental problem in computer vision.
It has many important applications.
For instance, Monocular Depth Estimation (MDE) is important for synthetic object insertion in computer graphics \cite{Karsch:2014:ASI:2631978.2602146}, grasping in robotics \cite{doi:10.1177/0278364914549607} and safety in self-driving cars. 
Face Normal Estimation can help in face image editing applications such as relighting \cite{SfSNet,Face_Relighting,DPR}.
However, it is extremely hard to annotate real data for these regression tasks.
Synthetic data and their ground truth labels, on the other hand, are easy to generate and are often used to compensate for the lack of labels in real data. 
Deep models trained on synthetic data, unfortunately, usually perform poorly on real data due to the domain gap between synthetic and real distributions.
To deal with this problem, several research studies \cite{adaDepth,T2NET,GASDA,Amir2018} have proposed unsupervised domain adaptation methods to take advantage of synthetic data by mapping it into the real domain or vice versa, either at the feature level or image level. 
However, mapping examples from one domain to another domain itself is a challenging problem that can limit performance.

We observe that finding such a mapping solves an unnecessarily difficult problem.  To train a regressor that applies to both real and synthetic domains, it is only necessary that we map both to a new representation that contains the task-relevant information present in both domains, in a common form.
The mapping need not alter properties of the original domain that are irrelevant to the task since the regressor will learn to ignore them regardless. 
%

To see this, we consider a simplified model of our problem. We suppose that real and synthetic images are formed by two components: domain agnostic (which has semantic information shared across synthetic and real, and is denoted as $I$) and domain specific.
%
We further assume that domain specific information has two sub-components: domain specific information unrelated to the primary task (denoted as $\delta_s'$ and $\delta_r'$ for synthetic and real images respectively) and domain specific information related to the primary task ($\delta_s$, $\delta_r$).
So real and synthetic images can be represented as: $x_r = f(I, \delta_r, \delta_r')$ and $x_s=f(I, \delta_s, \delta_s')$ respectively.

\begin{figure}
\centering
  \includegraphics[width=\linewidth]{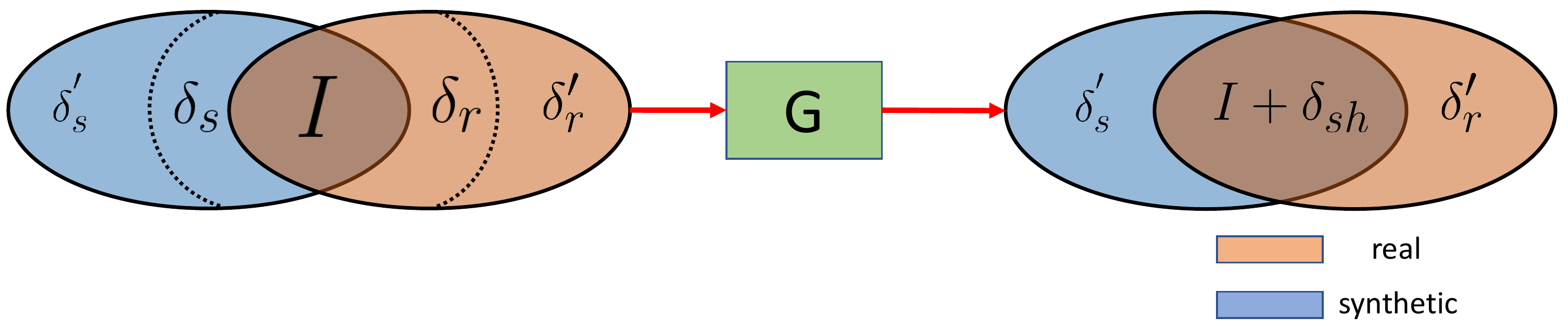}
    \caption{We propose to reduce the domain gap between synthetic and real by mapping the corresponding domain specific information related to the primary task $(\delta_s, \delta_r)$ into shared information $\delta_{sh}$, preserving everything else. }
\label{fig:SharinGAN-intro1}
\end{figure}

We believe the domain gap between $\{\delta_s$ and $\delta_r\}$ can affect the training of the primary network, which learns to expect information that is not always present.
The domain gap between $\{\delta_s'$ and $\delta_r'\}$, on the other hand, can be bypassed by the primary network since it does not hold information needed for the primary task.
For example, in real face images, information such as the color and texture of the hair is unrelated to the task of estimating face normals but is discriminative enough to distinguish real from synthetic faces. This can be regarded as domain specific information unrelated to the primary task i.e., $\delta_r'$.
%
%
On the other hand, shadows in the real and synthetic images, due to the limitations of the rendering engine, may have different appearances but may contain depth cues that are related to the primary task of MDE in both domains.
%
The simplest strategy, then, for combining real and synthetic data is to map $\delta_s$ and $\delta_r$ to a shared representation, $\delta_{sh}$, while not modifying $\delta'_s$ and $\delta'_r$ as shown in Figure \ref{fig:SharinGAN-intro1}.

Recent research studies show that a shared network for synthetic and real data can help reduce the discrepancy between images in different domains.
For instance, \cite{SfSNet} achieved state-of-the-art results in face normal estimation by training a unified network for real and synthetic data. 
\cite{CoGAN} learned the joint distribution of multiple domain images by enforcing a weight-sharing constraint for different generative networks.
Inspired by these research studies, we define a unified mapping function $G$, which is called SharinGAN, to reduce the domain gap between real and synthetic images.

Different from existing research studies, our $G$ is trained so that minimum domain specific information is removed.
This is achieved by pre-training $G$ as an auto-encoder on real and synthetic data, i.e., initializing $G$ as an identity function. Then $G$ is trained end-to-end with reconstruction loss in an adversarial framework, along with a network that solves the primary task, further pushing $G$ to map information relevant to the task to a shared domain.

As a result, a successfully trained $G$ will learn to reduce the domain gap existing in $\delta_s$ and $\delta_r$, mapping them into a shared domain $\delta_{sh}$.  $G$ will leave $I$ unchanged.   $\delta_s'$ and $\delta_r'$ can be left relatively unchanged when it is difficult to map them to a common representation.
Mathematically, $G(x_s) = f(I, \delta_{sh}, \delta_s')$ and $G(x_r) = f(I, \delta_{sh}, \delta_r')$. If successful, $G$ will map synthetic and real images to images that may look quite different to the eye, but the primary task network will extract the same information from both.

We apply our method to unsupervised monocular depth estimation using virtual KITTI (vKITTI) \cite{vKITTI} and KITTI \cite{KITTI} as synthetic and real datasets respectively.
Our method reduces the absolute error in the KITTI eigen test split and the test set of Make3D \cite{make3D} by $23.77 \%$ and $6.45 \%$ respectively compared with the state-of-the-art method \cite{GASDA}.
Additionally, our proposed method improves over SfSNet \cite{SfSNet} on face normal estimation.  It yields an accuracy boost of nearly $4.3 \%$ for normal prediction within $20^\circ$ $(Acc<20^\circ)$ of ground truth on the Photoface dataset \cite{Photoface}. Our code is available at \MYhref[magenta]{https://github.com/koutilya40192/SharinGAN}{https://github.com/koutilya40192/SharinGAN}.

\section{Related Work}
\begin{figure*}
    \centering
  \includegraphics[width=\linewidth]{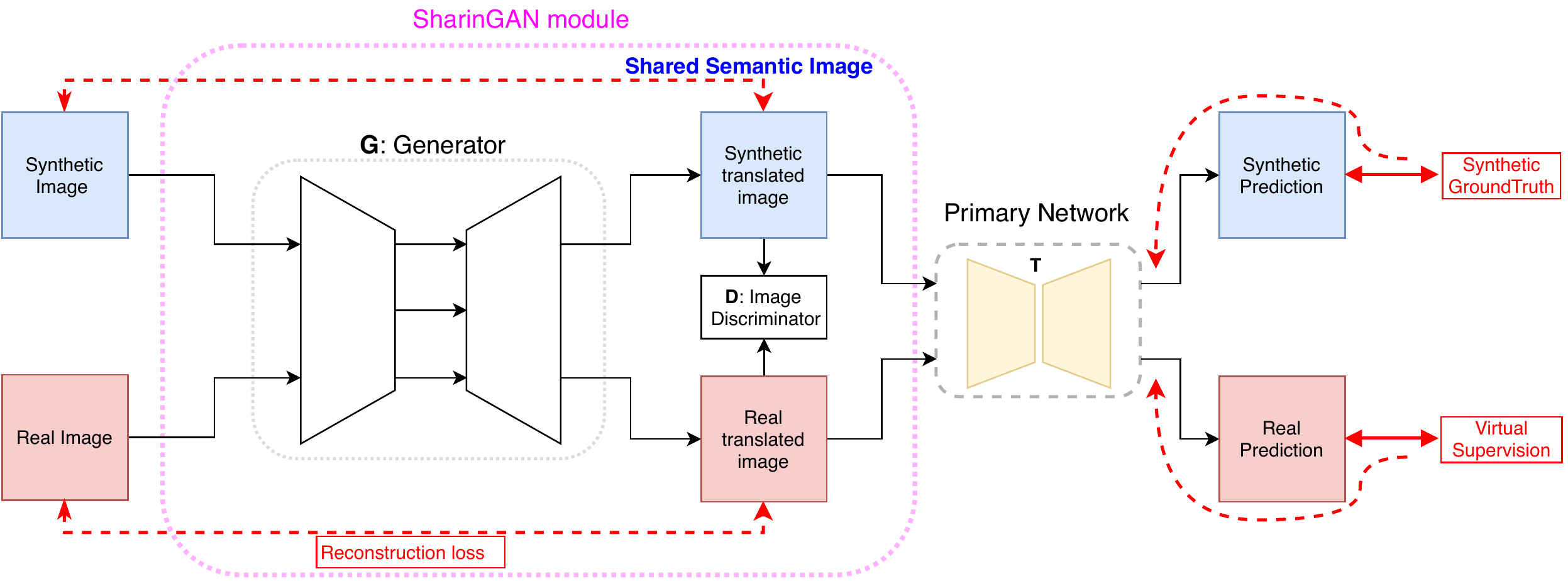}
    \caption{Overview of the model architecture. Red dashed arrows indicate the loss computations.}
\label{fig:SharinGAN}
\end{figure*}
\textbf{Monocular Depth Estimation} has long been an active area in computer vision.  Because this problem is ill-posed, learning-based methods have predominated in recent years.
Many early learning works applied Markov Random Fields (MRF) to infer the depth from a single image by modeling the relation between nearby regions \cite{Saxena2006,make3D,Liu_2014_CVPR}.
These methods, however, are time-consuming during inference and rely on manually defined features, which have limitations in performance.

More recent studies apply deep Convolutional Neural Networks (CNNs) \cite{Eigen2014,NYUv2,Fayao,Lei,Xu_2018_CVPR,repala2018dual,Qi_2018_CVPR,Roy_2016_CVPR} to monocular depth estimation.
Eigen \etal \cite{Eigen2014} first proposed a multi-scale deep CNN for depth estimation.
Following this work, \cite{NYUv2} proposed to apply CNNs to estimate depth, surface normal and semantic labels together.
\cite{Fayao} combined deep CNNs with a continuous CRF for monocular depth estimation.
One major drawback of these supervised learning-based methods is the requirement for a huge amount of annotated data, which is hard to obtain in reality.

With the emergence of large scale, high-quality synthetic data \cite{vKITTI}, using synthetic data to train a depth estimator network for real data became popular \cite{T2NET,GASDA}.
The biggest challenge for this task is the large domain gap between synthetic data and real data.
\cite{Amir2018} proposed to first train a depth prediction network using synthetic data.
A style transfer network is then trained to map real images to synthetic images in a cycle consistent manner \cite{CycleGAN2017}.
\cite{adaDepth} proposed to adapt the features of real images to the features of synthetic images by applying adversarial loss on latent features. 
A content congruent regularization is further proposed to avoid mode collapse.
T$^2$Net \cite{T2NET} trained a network that translates synthetic data into real at the image level and further trained a task network in this translated domain.
%
GASDA \cite{GASDA} proposed to train the network by incorporating epipolar geometry constraints for real data along with the ground truth labels for synthetic data.
All these methods try to align two domains by transferring one domain to another. 
Unlike these works, we propose a mapping function $G$, also called SharinGAN, to just align the domain specific information that affects the primary task, resulting in a minimum change in the images in both domains. 
We show that this makes learning the primary task network much easier and can help it focus on the useful information.

Self-supervised learning is another way to avoid collecting ground truth labels for monocular depth estimation. 
Such methods need monocular videos \cite{Zhou_2017_CVPR,Wang_2018_CVPR,AAAI_depth,Godard_2019_ICCV}, stereo pairs \cite{monodepth17,Mehta_2018,Poggi_2018,ma2018self}, or both\cite{Godard_2019_ICCV} for training. 
%
%
Our proposed method is complementary to these self-supervised methods, it does not require this additional data, but can use it when available.
%

\textbf{Face Geometry Estimation} is a sub-problem of inverse face rendering which is the key for many applications such as face image editing.
Conventional face geometry estimation methods are usually based on 3D Morphable Models (3DMM) \cite{3DMM}.
Recent studies demonstrate the effectiveness of deep CNNs for solving this problem \cite{Mofa,NeuralFace,3DMM_kyle,SfSNet,Luan_3DMM,tran2019towards,liu20193d}.
Thanks to the 3DMM, generating synthetic face images with ground truth geometry is easy. 
\cite{Mofa,NeuralFace,SfSNet} make use of synthetic face images with ground truth shape to help train a network for predicting face shape using real images.
Most of these works initially pre-train the network with synthetic data and then fine-tune it with a mix of real and synthetic data, either using no supervision or weak supervision, overlooking the domain gap between real and synthetic face images.
In this work, we show that by reducing the domain gap between real and synthetic data using our proposed method, face geometry can be better estimated.

\camready{\textbf{Domain Adaptation using GANs} There are many works \cite{Tzeng_2017_CVPR,Bousmalis_2017_CVPR,CoGAN,taigman2016unsupervised,shen2017wasserstein} that use a GAN framework to perform domain adaptation by mapping one domain into another via a supervised translation. However, most of these show performance on just toy datasets in a classification setting. We attempt to map both synthetic and real domains into a new shared domain that is learned during training and use this to solve complex problems of unsupervised geometry estimation. Moreover, we apply adversarial loss at the image level for our regression task, in contrast to some of the above previous works where domain invariant feature engineering sufficed for classification tasks.}

\section{Method}
To compensate for the lack of annotations for real data and to train a primary task network on easily available synthetic data, we propose SharinGAN to reduce the domain gap between synthetic and real. 
We aim to train a primary task network on a shared domain created by SharinGAN, which learns the mapping function $G:x_r \mapsto x_r^{sh}$ and  $G:x_s \mapsto x_s^{sh}$, where $x_k = f(I, \delta_k, \delta_k'); \thickspace x_k^{sh} = f(I, \delta_{sh}, \delta_k'); \thickspace k \in \{r,s\}$ as shown in Figure \ref{fig:SharinGAN-intro1}. 
$G$ allows the primary task network to train on a shared space that holds the information needed to do the primary task, making the network more applicable to real data during testing.

To achieve this, an adversarial loss is used to find the shared information, $\delta_{sh}$. 
This is done by minimizing the discrepancy in the distributions of $x_r^{sh}$ and $x_s^{sh}$. 
But at the same time, to preserve the domain agnostic information (shared semantic information $I$), we use reconstruction loss. 
Now, without a loss from the primary task network, $G$ might change the images so that they don't match the labels. 
To prevent that, we additionally use a primary task loss for both real and synthetic examples to guide the generator. 
It is important to note that both the translations from synthetic to real and vice versa are equally crucial for this symmetric setup to find a shared space. 
To facilitate that, we use a form of weak supervision we call virtual supervision. Some possible virtual supervisions include a prior on the input data or a constraint that can narrow the solution space for the primary task network \camready{(details discussed in \ref{vs}).}
%
%
For synthetic examples, we use the known labels.

Adversarial, Reconstruction and Primary task losses together train the generator and primary task network to align the domain specific information $\{\delta_s, \delta_r\}$ in both the domains into a shared space $\delta_{sh}$, preserving everything else.

\subsection{Framework}
In this work, we propose to train a generative network which is called SharinGAN, to reduce the domain gap between real and synthetic data so as to help to train the primary network.
Figure~\ref{fig:SharinGAN} shows the framework of our proposed method.
It contains a generative network $G$, a discriminator on image-level $D$ that embodies the SharinGAN module and a task network $T$ to perform the primary task.
The generative network $G$ takes either a synthetic image $x_s$ or real image $x_r$ as input and transforms it to $x_s^{sh}$ or $x_r^{sh}$ in an attempt to fool $D$.
Different from existing works that transfer images in one domain to another \cite{Amir2018,T2NET,GASDA}, our generative network $G$ tries to map the domain specific parts $\delta_s$ and $\delta_r$ of synthetic and real images to a shared space $\delta_{sh}$, leaving $\delta_s'$ and $\delta_r'$ unchanged.
As a result, our transformed synthetic and real images ($x_s^{sh}$ and $x_r^{sh}$) have fewer differences from $x_s$ and $x_r$.
Our task network $T$ then takes the transformed images $x_s^{sh}$ and $x_r^{sh}$ as input and predicts the geometry.
The generative network $G$ and task network $T$ are trained together in an end-to-end manner.

\subsection{Losses}
In this section, we describe the losses we use for the generative and task networks. 
\subsubsection{Losses for Generative Network}
We design a single generative network $G$ for synthetic and real data since sharing weights can help align distributions of different domains \cite{CoGAN}.
Moreover, existing research studies such as \cite{NeuralFace, SfSNet} also demonstrate that a unified framework works reasonably well on synthetic and real images.
In order to map $\delta_s$ and $\delta_r$ to a shared space $\delta_{sh}$, we apply adversarial loss \cite{GAN} at the image level.
More specifically, we use the Wasserstein discriminator \cite{WGAN} that uses the Earth-Mover's distance to minimize the discrepancy between the distributions for synthetic and real examples $\{G(x_s), G(x_r)\}$, i.e.:
%
%
\begin{eqnarray}
    L_{W}(D, G) = \mathbb{E}_{x_s} [D(G(x_s))] - \mathbb{E}_{x_r } [D(G(x_r))], \label{eq:gF}
\end{eqnarray}
$D$ is a discriminator and $G_e$ is the encoder part of the generator. 
Following \cite{WGANGP}, to overcome the problem of vanishing or exploding gradients due to the weight clipping proposed in \cite{WGAN}, a gradient penalty term is added for training the discriminator: 
\begin{eqnarray}
 L_{gp}(D) &=& (|| \nabla_{\hat{h}} D(\hat{h}) ||_2 - 1)^2 
\end{eqnarray}
Our overall adversarial loss is then defined as:
\begin{eqnarray}
L_{adv} &=& L_{W}(D, G) - \lambda L_{gp}(D)
\end{eqnarray}
where $\lambda$ is chosen to be $10$ while training the discriminator and $0$ while training the generator.

Without any constraints, the adversarial loss may learn to remove all domain specific parts $\delta$ and $\delta'$ or even some of the domain agnostic part $I$ in order to fool the discriminator.
This may lead to loss of geometric information, which can degrade the performance of the primary task network $T$.
To avoid this, we propose to use the self-regularization loss similar to \cite{SimGAN} to force the transformed image to keep as much information as possible:
\begin{equation}
    L_{r} = ||G(x_s) - x_s||_2^2 + ||G(x_r) - x_r||_2^2.
\end{equation}

\begin{table*}
    \centering
    \begin{adjustbox}{width=\linewidth}
    \begin{tabular}{|l||c|c|c||c|c|c|c|c|c|c|}
    \hline
    \multirow{2}{*}{Method} & \multirow{2}{*}{Supervised} & \multirow{2}{*}{Dataset} & \multirow{2}{*}{Cap} & \multicolumn{4}{|c|}{Error Metrics, lower is better} & \multicolumn{3}{|c|}{Accuracy Metrics, higher is better}\\
    \cline{5-11}
    & & & & Abs Rel & Sq Rel & RMSE & RMSE log &  $\delta < 1.25$ & $\delta < 1.25^2$ & $\delta < 1.25^3$\\
    
    \hline
        Eigen \etal \cite{Eigen2014} & Yes & K & 80m & 0.203 & 1.548 & 6.307 & 0.282 & 0.702 & 0.890 & 0.958\\
        Liu \etal \cite{Fayao} & Yes & K & 80m & 0.202 & 1.614 & 6.523 & 0.275 & 0.678 & 0.895 & 0.965\\
        \hline
        All synthetic (baseline) & No & S & 80m & 0.253 & 2.303 & 6.953 & 0.328 & 0.635 & 0.856 & 0.937\\
        All real (baseline) & No & K & 80m & 0.158 & 1.151 & 5.285 & 0.238 & 0.811 & 0.934 & 0.970\\
        \hline
        \rowcolor{lightgray} GASDA \cite{GASDA} & No & K+S & 80m & 0.149 & 1.003 & \textbf{4.995} & 0.227 & 0.824 & 0.941 & 0.973\\
        \rowcolor{lightgray} SharinGAN (proposed) & No & K+S & 80m &  \textbf{0.116} & \textbf{0.939} & 5.068 & \textbf{0.203} & \textbf{0.850} & \textbf{0.948} & \textbf{0.978}\\
        \hline
        \hline
        Kuznietsov \etal \cite{Kuznietsov_2017_CVPR} & Yes & K & 50m & 0.117 & 0.597 & 3.531 & 0.183 & 0.861 & 0.964 & 0.989\\
        Garg \etal \cite{Garg_2016} & No & K & 50m & 0.169 & 1.080 & 5.104 & 0.273 & 0.740 & 0.904 & 0.962 \\
        Godard \etal \cite{monodepth17} & No & K & 50m & 0.140 & 0.976 & 4.471 & 0.232 & 0.818 & 0.931 & 0.969 \\
        \hline
        All synthetic (baseline) & No & S & 50m & 0.244 & 1.771 & 5.354 & 0.313 & 0.647 & 0.866 & 0.943\\
        All real (baseline) & No & K & 50m & 0.151 & 0.856 & 4.043 & 0.227 & 0.824 & 0.940 & 0.973\\
         \hline
         \rowcolor{lightgray} Kundu \etal \cite{adaDepth} & No & K+S & 50m & 0.203 & 1.734 & 6.251 & 0.284 &0.687 &0.899 & 0.958 \\
         \rowcolor{lightgray} T2Net \cite{T2NET} & No & K+S & 50m & 0.168 & 1.199 & 4.674 & 0.243 & 0.772 & 0.912 & 0.966\\
         \rowcolor{lightgray} GASDA \cite{GASDA} & No & K+S & 50m & 0.143 & 0.756 & 3.846 & 0.217 & 0.836 & 0.946 & 0.976\\
         \rowcolor{lightgray} SharinGAN (proposed) & No & K+S & 50m & \textbf{0.109} & \textbf{0.673} & \textbf{3.77} & \textbf{0.190} & \textbf{0.864} & \textbf{0.954} & \textbf{0.981}\\
    \hline
    \end{tabular}
    \end{adjustbox}
    \caption{MDE Results on eigen test split of KITTI dataset \cite{Eigen2014} . For the training data, K: KITTI dataset and S: vKITTI dataset. Methods highlighted in light gray, use domain adaptation techniques and the non-highlighted rows correspond to supervised methods.}
    \vspace{-5mm}
    \label{tab:mde_results}
\end{table*}

\subsubsection{Losses for the Task Network}\label{vs}
The task network takes transformed synthetic or real images as input and predicts geometric information.
Since the ground truth labels for synthetic data are available, we apply a supervised loss using these ground truth labels.
%
For real images, domain specific losses or regularizations are applied \camready{as a form of virtual supervision} for training according to the task. 
We apply our proposed SharinGAN to two tasks: monocular depth estimation \camready{(MDE)} and face normal estimation \camready{(FNE)}. 
\camready{For MDE, we use the combination of depth smoothness and geometric consistency losses used in GASDA \cite{GASDA} as the virtual supervision. For FNE however, for virtual supervision we use the pseudo supervision used in SfSNet \cite{SfSNet}. We use the term \say{virtual supervision} to summarize these two losses as a kind of weak supervision on the real examples.}

\textbf{Monocular Depth Estimation}.
To make use of ground truth labels for synthetic data, we apply $L_1$ loss for predicted synthetic depth images:
\begin{eqnarray}
L_1 = ||\hat{y}_s - y_s^*||_1
\end{eqnarray}
where $\hat{y}_s$ is the predicted synthetic depth map and $y_s^*$ is its corresponding ground truth.
Following \cite{GASDA}, we apply smoothness loss on depth $L_{DS}$ to encourage it to be consistent with local homogeneous regions. Geometric consistency loss $L_{GC}$ is applied so that the task network can learn the physical geometric structure through epipolar constraints.
$L_{DS}$ and $L_{GC}$ are defined as:
\begin{equation}
L_{DS} =e^{-\nabla x_r}||\nabla \hat{y_r}||     \label{eq:DS}
\end{equation}
\begin{equation}
L_{GC} = \eta \frac{1-SSIM(x_{r},x_{rr}')}{2} + \mu ||x_{r} - x_{rr}' || \label{eq:GC},
\end{equation}
$\hat{y}_r$ represents the predicted depth for the real image and $\nabla$ represents the first derivative.
%
%
%
%
$x_r$ is the left image in the KITTI dataset \cite{KITTI}. 
$x_{rr}'$ is the inverse warped image from the right counterpart of $x_r$ based on the predicted depth $\hat{y}_r$.
The KITTI dataset\cite{KITTI} provides the camera focal length and the baseline distance between the cameras. 
Similar to \cite{GASDA}, we set $\eta$ as 0.85 and $\mu$ as 0.15 in our experiments.
%
The overall loss for the task network is defined as:
\begin{equation}
L_T = \beta_1L_{DS} + \beta_2L_1  + \beta_3L_{GC},
\label{eq:PT}
\end{equation}
where $\beta_1 = 0.01, \beta_2 = \beta3 = 100.$

\textbf{Face Normal Estimation.} 
SfSnet \cite{SfSNet} currently achieves the best performance on face normal estimation.
We thus follow its setup for face normal estimation and apply ``SfS-supervision'' for both synthetic and real images during training.
\begin{eqnarray}
L_{T} = \lambda_{recon}L_{recon} + \lambda_{N}L_{N} + \lambda_{A}L_{A} + \lambda_{Light}L_{Light},
\end{eqnarray}
where $L_{recon}$, $L_{N}$ and $L_{A}$ are $L_1$ losses on the reconstructed image, normal and albedo, whereas $L_{light}$ is the L2 loss over the 27 dimensional spherical harmonic coefficients.
The supervision for real images is from the ``pseudo labels'', obtained by applying a pre-trained task network on real images.
Please refer to \cite{SfSNet} for more details.

\subsection{Overall loss}
The overall loss used to train our geometry estimation pipeline is then defined as:
\begin{equation}
L = \alpha_1 L_{adv} + \alpha_2 L_{r} + \alpha_3 L_{T}. \label{eq:overall}
\end{equation}
where $(\alpha_1, \alpha_2, \alpha_3) = (1,10,1)$ for monocular depth estimation task and $(\alpha_1, \alpha_2, \alpha_3) = (1,10,0.1)$ for face normal estimation task.




\section{Experiments}
We apply our proposed SharinGAN to monocular depth estimation and face normal estimation.
We discuss the details of the experiments in this section.
\begin{figure*}
\begin{subfigure}[t]{\linewidth}
    \begin{subfigure}[t]{0.25\linewidth}
        \includegraphics[width=\linewidth]{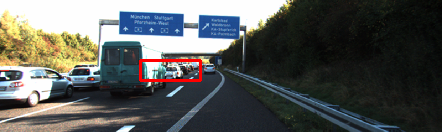}
        \includegraphics[width=\linewidth]{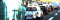}
    \end{subfigure}%
    ~
    \begin{subfigure}[t]{0.25\linewidth}
        \includegraphics[width=\linewidth]{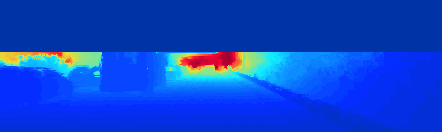}
        \includegraphics[width=\linewidth]{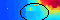}
    \end{subfigure}%
    ~
    \begin{subfigure}[t]{0.25\linewidth}
        \includegraphics[width=\linewidth]{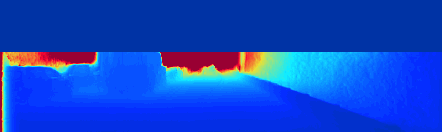}
        \includegraphics[width=\linewidth]{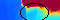}
    \end{subfigure}%
    ~
    \begin{subfigure}[t]{0.25\linewidth}
        \includegraphics[width=\linewidth]{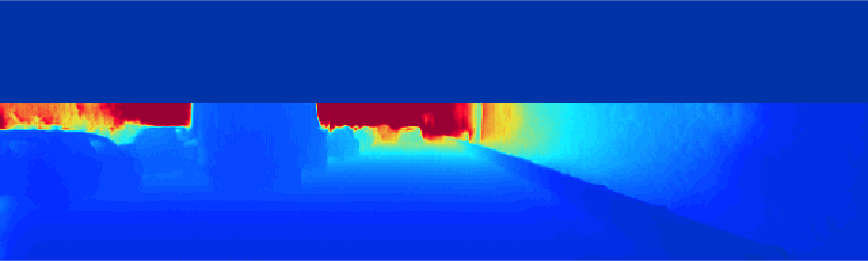}
        \includegraphics[width=\linewidth]{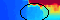}
    \end{subfigure}%
    \caption{First row from left to right: real image, ground truth depth map, depth map by GASDA \cite{GASDA} and depth map by SharinGAN. The second row shows the corresponding region in the red box of the first row. The depth of the faraway car is better estimated by SharinGAN than GASDA.}
    \vspace*{4mm}
\end{subfigure}%

\begin{subfigure}[t]{\linewidth}
    \begin{subfigure}[t]{0.25\linewidth}
        \includegraphics[width=\linewidth]{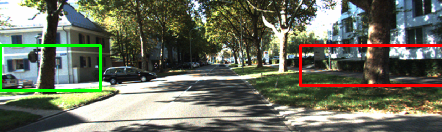}
        \includegraphics[width=\linewidth]{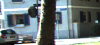}
        \includegraphics[width=\linewidth]{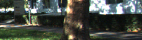}
    \end{subfigure}%
    ~
    \begin{subfigure}[t]{0.25\linewidth}
        \includegraphics[width=\linewidth]{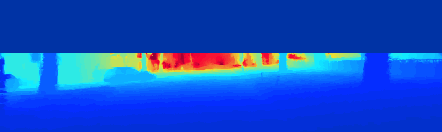}
        \includegraphics[width=\linewidth]{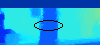}
        \includegraphics[width=\linewidth]{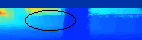}
    \end{subfigure}%
    ~
    \begin{subfigure}[t]{0.25\linewidth}
        \includegraphics[width=\linewidth]{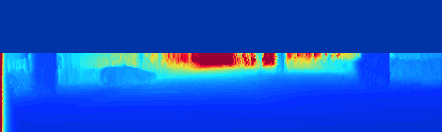}
        \includegraphics[width=\linewidth]{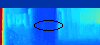}
        \includegraphics[width=\linewidth]{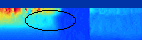}
    \end{subfigure}%
    ~
    \begin{subfigure}[t]{0.25\linewidth}
        \includegraphics[width=\linewidth]{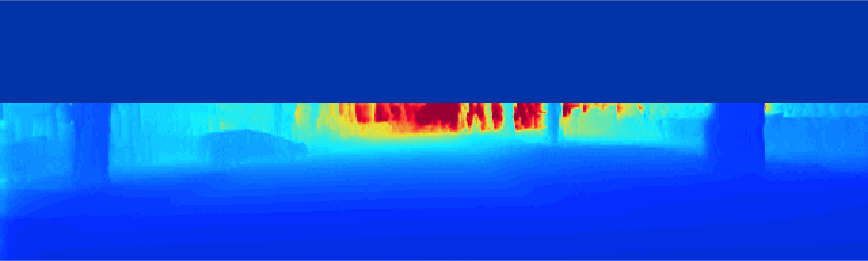}
        \includegraphics[width=\linewidth]{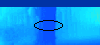}
        \includegraphics[width=\linewidth]{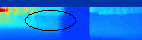}
    \end{subfigure}%
    \caption{First row from left to right: real image, ground truth depth map, depth map by GASDA \cite{GASDA} and depth map by SharinGAN. The second and third row shows the corresponding region in the green and red box of the first row. The depth of the tree to the left (green) and shrubs behind the tree in the right are better estimated by SharinGAN.}
    \vspace*{4mm}
\end{subfigure}%

\begin{subfigure}[t]{\linewidth}
    \begin{subfigure}[t]{0.25\linewidth}
        \includegraphics[width=\linewidth]{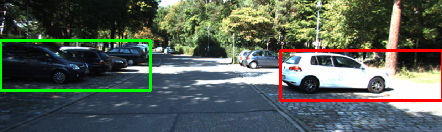}
        \includegraphics[width=\linewidth]{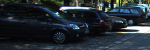}
        \includegraphics[width=\linewidth]{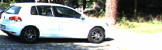}
    \end{subfigure}%
    ~
    \begin{subfigure}[t]{0.25\linewidth}
        \includegraphics[width=\linewidth]{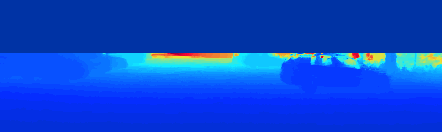}
        \includegraphics[width=\linewidth]{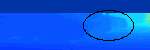}
        \includegraphics[width=\linewidth]{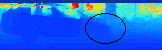}
    \end{subfigure}%
    ~
    \begin{subfigure}[t]{0.25\linewidth}
        \includegraphics[width=\linewidth]{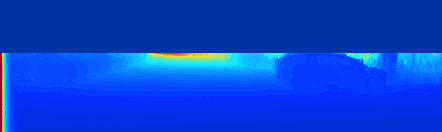}
        \includegraphics[width=\linewidth]{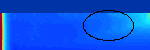}
        \includegraphics[width=\linewidth]{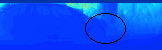}
    \end{subfigure}%
    ~
    \begin{subfigure}[t]{0.25\linewidth}
        \includegraphics[width=\linewidth]{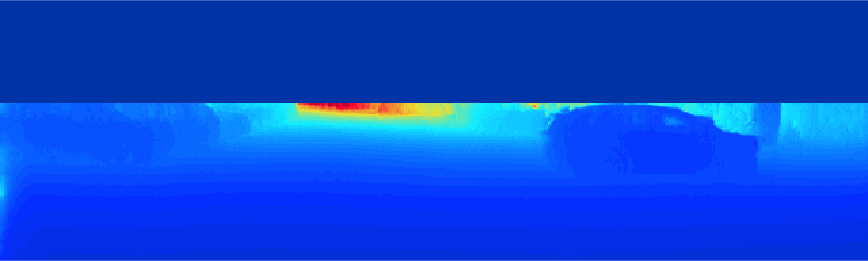}
        \includegraphics[width=\linewidth]{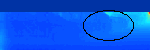}
        \includegraphics[width=\linewidth]{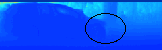}
    \end{subfigure}%
    \caption{First row from left to right: real image, ground truth depth map, depth map by GASDA \cite{GASDA} and depth map by SharinGAN. The second and third row shows the corresponding region in the green and red box of the first row. The \camready{boundaries and the} depth of the cars are better estimated by SharinGAN.}
\end{subfigure}%

\caption{Qualitative comparisons of SharinGAN with GASDA \cite{GASDA}. Ground truth (GT) has been interpolated for visualization. We mask out the top regions where ground truth depth is not available for visualization purposes. Note that in addition to various other aspects mentioned above, we are also able to remove the boundary artifacts present in the depth maps of GASDA.}
\vspace{-1mm}
\label{fig:MDE_SharinGAN_visualizations}
\end{figure*}

\subsection{Monocular Depth Estimation}
\textbf{Datasets} Following \cite{GASDA}, we use vKITTI \cite{vKITTI} and KITTI \cite{KITTI} as synthetic and real datasets to train our network.
vKITTI contains $21,260$ image-depth pairs, which are all used for training.
KITTI \cite{KITTI} provides $42,382$ stereo pairs, among which, $22,600$ images are used for training and $888$ are used for validation as suggested by \cite{GASDA}.

\textbf{Implementation details}
We use a generator $G$ and a primary task network $T$, whose architectures are identical to \cite{GASDA}.
We pre-train the generative network $G$ on both synthetic and real data using reconstruction loss $L_r$.
This results in an identity mapping that can help $G$ to keep as much of the input image's geometry information as possible.
Our task network is pre-trained using synthetic data with supervision.
$G$ and $T$ are then trained end to end using Equation~\ref{eq:overall} for 150,000 iterations with a batch size of 2, by using an Adam optimizer with a learning rate of $1e-5$.
The best model is selected based on the validation set of KITTI.

\begin{figure*}
\begin{subfigure}{0.49\linewidth}
    \centering
    \begin{subfigure}{0.33\linewidth}{
        \includegraphics[width=\linewidth]{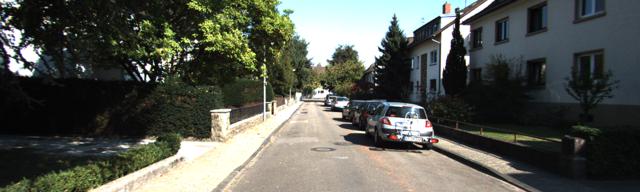}}
    \end{subfigure}%
    ~
    \begin{subfigure}{0.33\linewidth}{
        \includegraphics[width=\linewidth]{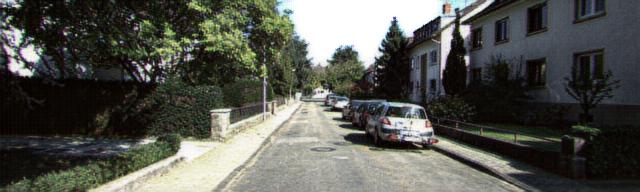}}
    \end{subfigure}%
    ~
    \begin{subfigure}{0.33\linewidth}{
        \includegraphics[width=\linewidth]{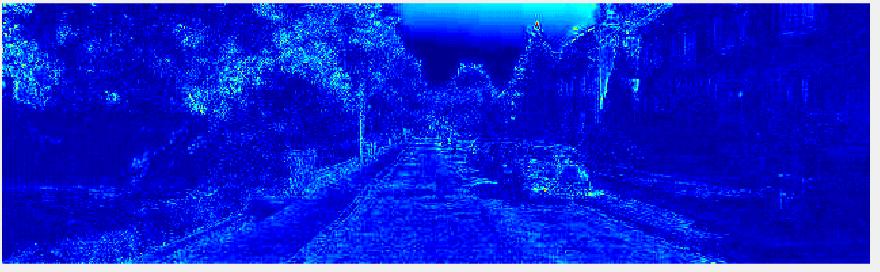}}
    \end{subfigure}%
    
    \begin{subfigure}{0.33\linewidth}{
        \includegraphics[width=\linewidth]{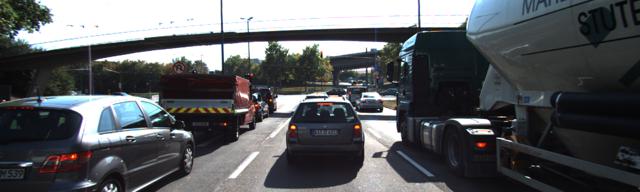}}
        \caption{$x_r$}
    \end{subfigure}%
    ~
    \begin{subfigure}{0.33\linewidth}{
        \includegraphics[width=\linewidth]{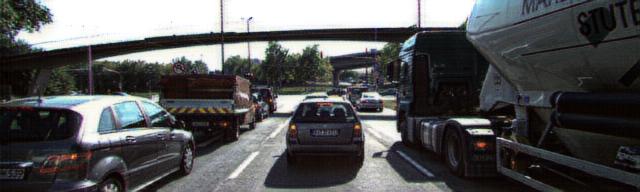}}
        \caption{$x_r^{sh} = G(x_r)$}
    \end{subfigure}%
    ~
    \begin{subfigure}{0.33\linewidth}{
        \includegraphics[width=\linewidth]{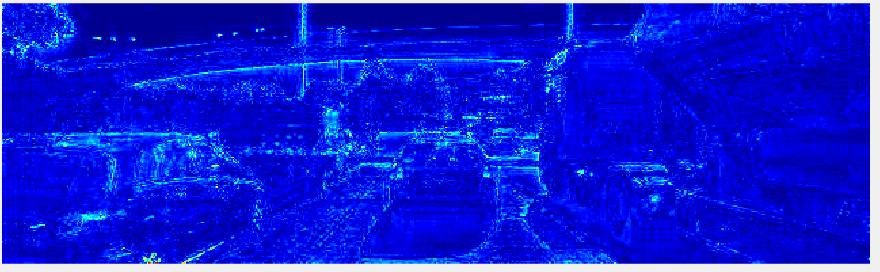}}
        \caption{$|x_r - x_r^{sh}|$}
    \end{subfigure}%
\end{subfigure}%
\hspace{0.02\textwidth}
~
\begin{subfigure}{0.49\linewidth}
    \centering
    \begin{subfigure}{0.33\linewidth}{
        \includegraphics[width=\linewidth]{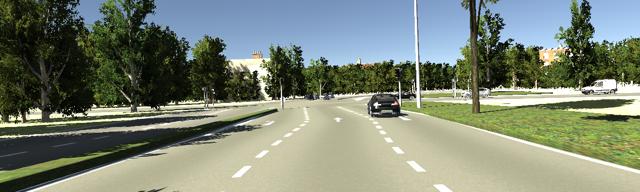}}
    \end{subfigure}%
    ~
    \begin{subfigure}{0.33\linewidth}{
        \includegraphics[width=\linewidth]{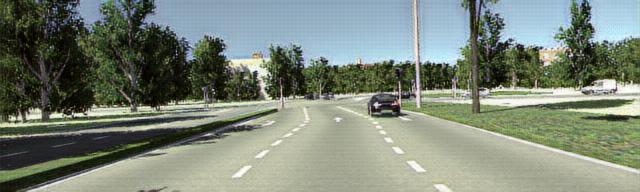}}
    \end{subfigure}%
    ~
    \begin{subfigure}{0.33\linewidth}{
        \includegraphics[width=\linewidth]{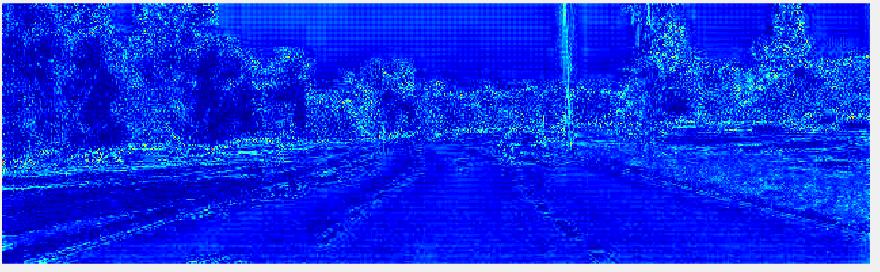}}
    \end{subfigure}%
    
    \begin{subfigure}{0.33\linewidth}{
        \includegraphics[width=\linewidth]{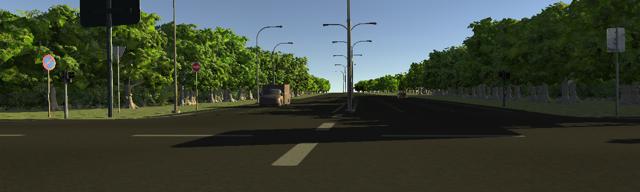}}
        \caption{$x_s$}
    \end{subfigure}%
    ~
    \begin{subfigure}{0.33\linewidth}{
        \includegraphics[width=\linewidth]{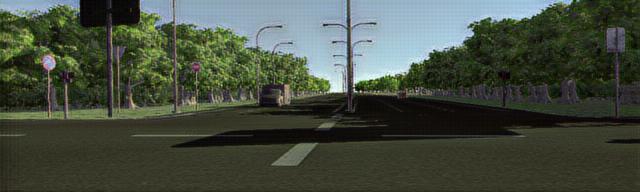}}
        \caption{$x_s^{sh} = G(x_s)$}
    \end{subfigure}%
    ~
    \begin{subfigure}{0.33\linewidth}{
        \includegraphics[width=\linewidth]{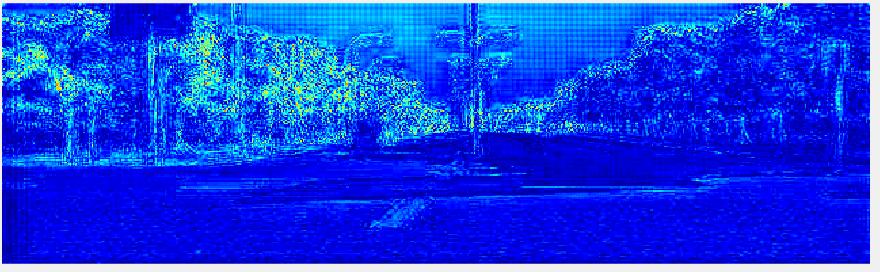}}
        \caption{$|x_s - x_s^{sh}|$}
    \end{subfigure}%
\end{subfigure}%
\caption{(a), (b) and (c) show real image $x_r$, translated real image $x_r^{sh}$ and their difference $|x_r - x_r^{sh}|$ respectively. (d), (e) and (f) show synthetic image $x_s$, translated synthetic image $x_s^{sh}$ and their difference $|x_s - x_s^{sh}|$ respectively.}
\label{fig:MDE_SharinGAN}
\end{figure*}

\begin{figure*}[!t]
    \begin{subfigure}{0.25\linewidth}{
        \includegraphics[width=\linewidth]{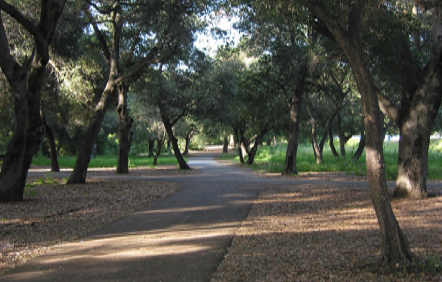}
        \includegraphics[width=\linewidth]{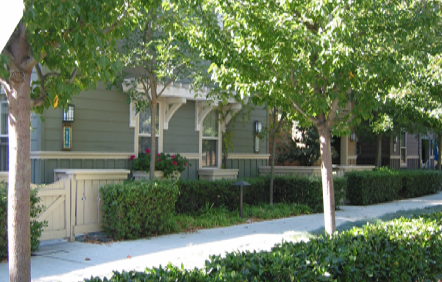}}
        \caption{Input Image}
    \end{subfigure}%
    ~
    \begin{subfigure}{0.25\linewidth}{
        \includegraphics[width=\linewidth]{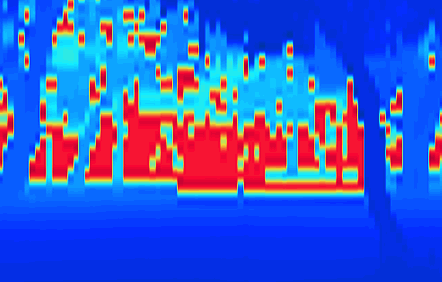}
        \includegraphics[width=\linewidth]{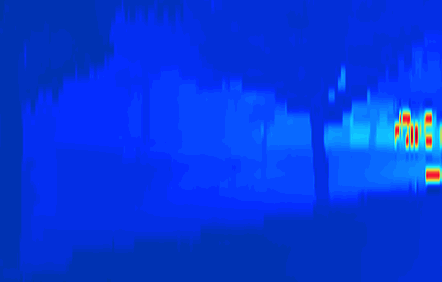}}
        \caption{Ground Truth}
    \end{subfigure}%
    ~
    \begin{subfigure}{0.25\linewidth}{
        \includegraphics[width=\linewidth]{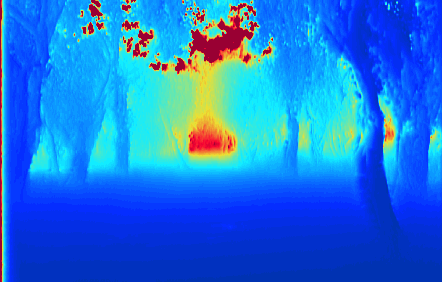}
        \includegraphics[width=\linewidth]{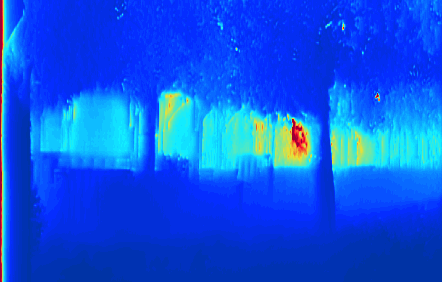}}
        \caption{GASDA\cite{GASDA}}
    \end{subfigure}%
    ~
    \begin{subfigure}{0.25\linewidth}{
        \includegraphics[width=\linewidth]{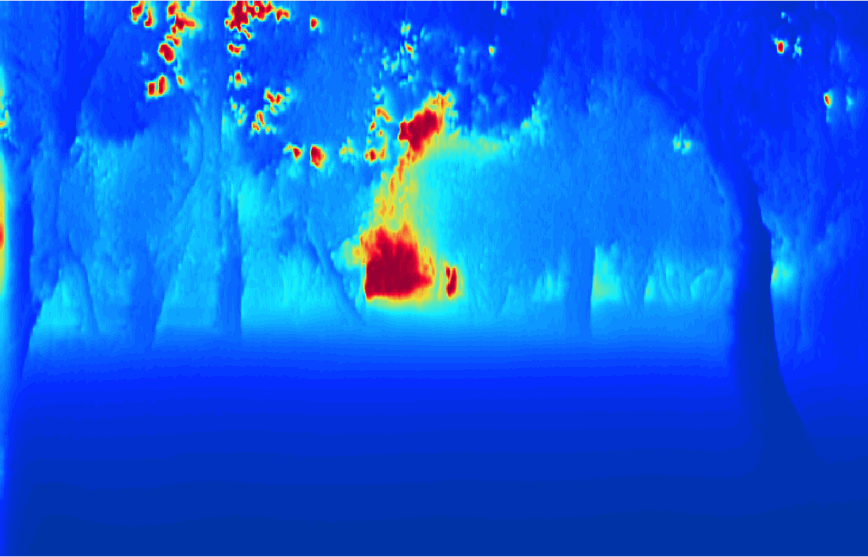}
        \includegraphics[width=\linewidth]{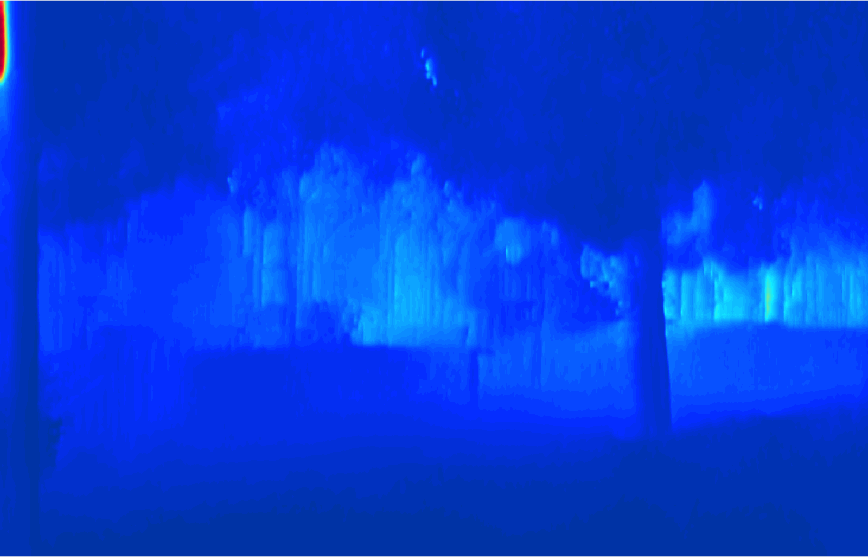}}
        \caption{SharinGAN}
    \end{subfigure}%
\caption{Qualitative results on the test set of the Make3D dataset \cite{make3D}. In the top row, some far tree structures that are missing in the depth map predicted by GASDA were better captured on using the SharinGAN module. For the bottom row, GASDA wrongly predicts the depth map of the houses behind the trees to be far, which is correctly captured by the SharinGAN.}
\label{fig:MDE_SharinGAN_make3d}
\end{figure*}

\textbf{Results}
Table \ref{tab:mde_results} shows the quantitative results on the eigen test split of the KITTI dataset for different methods on the MDE task. 
The proposed method outperforms the previous unsupervised domain adaptation methods for MDE \cite{GASDA,T2NET} on almost all the metrics.
Especially, compared with \cite{GASDA}, we reduce the absolute error by $19.7\%$ and $21.0\%$ on 80m cap and 50m cap settings respectively.
Moreover, the performance of our method is much closer to the methods in a supervised setting \cite{Eigen2014,Fayao,Kuznietsov_2017_CVPR}, which was trained on the real KITTI dataset with ground truth depth labels. 
Figure~\ref{fig:MDE_SharinGAN_visualizations} visually compares the predicted depth map from the proposed method with \cite{GASDA}.
We show three typical examples: near distance, medium distance, and far distance.
It shows that our proposed method performs much better for predicting depth at details.
For instance, our predicted depth map can better preserve the shape of the car (Figure~\ref{fig:MDE_SharinGAN_visualizations} (a) and (c)) and the structure of the tree and the building behind it (Figure~\ref{fig:MDE_SharinGAN_visualizations} (b)). 
This shows the advantage of our proposed SharinGAN compared with \cite{GASDA}.
\cite{GASDA} learns to transfer real images to the synthetic domain and vice versa, which solves a much harder problem compared with SharinGAN, which removes a minimum of domain specific information.
As a result, the quality of the transformation for \cite{GASDA} may not be as good as the proposed method.
Moreover, the unsupervised transformation cannot guarantee to keep the geometry information unchanged.

To understand how our generative network $G$ works, we show some examples of synthetic and real images, their transformed versions, and the difference images in Figure~\ref{fig:MDE_SharinGAN}.
This shows that $G$ mainly operates on edges.
Since depth maps are mostly discontinuous at edges, they provide important cues for the geometry of the scene.
On the other hand, due to the difference between the geometry and material of objects around the edges, the rendering algorithm may find it hard to render realistic edges compared with other parts of the scene.
As a result, most of the domain specific information related to geometry lies in the edges, on which SharinGAN correctly focuses.

\subsubsection{Generalization to Make3D}
To demonstrate the generalization ability of the proposed method, we test our trained model on Make3D \cite{make3D}.
Note that we do not fine-tune our model using the data from Make3D.
Table~\ref{tab:mde_results_make3d} shows the quantitative results of our method, which outperforms existing state-of-the-art methods by a large margin. 
\begin{table}[!h]
    
    \begin{adjustbox}{width=\linewidth}
    \centering
    \begin{tabular}{|l||c||c|c|c|}
    \hline
    \multirow{2}{*}{Method} & \multirow{2}{*}{Trained} & \multicolumn{3}{|c|}{Error Metrics, lower is better}\\
    \cline{3-5}
    & & Abs Rel & Sq Rel & RMSE\\
    
    \hline
        Karsh et al. \cite{Karsch} & Yes & 0.398 & 4.723 & 7.801\\
        Laina et al. \cite{Laina} & Yes & 0.198 & 1.665 & 5.461\\
        Kundu et al. \cite{adaDepth} & Yes & 0.452 & 5.71 & 9.559\\
        \hline
        \hline
        Goddard et al. \cite{Godard_2017_CVPR} & No & 0.505 & 10.172 & 10.936\\
        Kundu et al. \cite{adaDepth} & No & 0.647 & 12.341 & 11.567\\
        Atapour et al. \cite{Amir2018} & No & 0.423 & 9.343 & 9.002\\
        T2Net \cite{T2NET} & No & 0.508 & 6.589 & 8.935\\
        GASDA \cite{GASDA} & No & 0.403 & 6.709 & 10.424\\
        SharinGAN (proposed) & No & \textbf{0.377} & \textbf{4.900} & \textbf{8.388}\\
    \hline
    \end{tabular}
    \end{adjustbox}
    \caption{MDE results on Make3D dataset \cite{make3D}. Trained indicates whether the model is trained on Make3D or not. Errors are computed for depths less than 70m in a central image crop \cite{Godard_2017_CVPR}. It can be concluded that our proposed method generalized better to an unseen dataset.}
    \label{tab:mde_results_make3d}
\end{table}
Moreover, the performance of SharinGAN is more comparable to the supervised methods.
We further visually compare the proposed method with GASDA \cite{GASDA} in Figure~\ref{fig:MDE_SharinGAN_make3d}.
It is clear that the proposed depth map captures more details in the input images, reflecting more accurate depth prediction.


\begin{table*}[ht]
    \begin{adjustbox}{width=\linewidth}
    \centering
    \begin{tabular}{|c|c||c||c|c|c|c|c|c|c|}
    \hline
    \multicolumn{2}{|c||}{Components} & \multirow{2}{*}{Cap} & \multicolumn{4}{|c|}{Error Metrics, lower is better} & \multicolumn{3}{|c|}{Accuracy Metrics, higher is better}\\
    \cline{1-2} \cline{4-10}
    SharinGAN & Reconstruction loss && Abs Rel & Sq Rel & RMSE & RMSE log &  $\delta < 1.25$ & $\delta < 1.25^2$ & $\delta < 1.25^3$\\
    \hline
        x & x & 50m & 0.137 & 0.804 & 4.12 & 0.210 & 0.816 & 0.940 & 0.978\\
        \checkmark & x & 50m & 0.1113 & \textbf{0.6705} & 3.80 & 0.192 & 0.861 & 0.954 & 0.980\\
        \checkmark & \checkmark & 50m & \textbf{0.109} & 0.673 & \textbf{3.77} & \textbf{0.190} & \textbf{0.864} & \textbf{0.954} & \textbf{0.981}\\
    \hline
    \end{tabular}
    \end{adjustbox}
    \caption{Ablation study for monocular depth estimation to understand the role of the SharinGAN module and Reconstruction loss. We need both to get the best performance for this task.}
    \vspace{-5mm}
    \label{tab:ablation_sharingan_training}
\end{table*}

\subsection{Face Normal Estimation}
\textbf{Datasets} We use the synthetic data provided by \cite{SfSNet} and CelebA \cite{celebA} as real data to train the SharinGAN for face normal estimation similar to \cite{SfSNet}.
Our trained model is then evaluated on the Photoface dataset \cite{Photoface}.

\textbf{Implementation details} We use the RBDN network \cite{Santhanam_2017_CVPR} as our generator and SfSNet \cite{SfSNet} as the primary task network. Similar to before, we pre-train the Generator on both synthetic and real data using reconstruction loss and pre-train the primary task network on just synthetic data in a supervised manner. Then, we train $G$ and $T$ end-to-end using the overall loss (\ref{eq:overall}) for 120,000 iterations. We use a batch size of 16 and a learning rate of $1e-4$. The best model is selected based on the validation set of Photoface\cite {Photoface}.
\begin{table}[h]
\begin{adjustbox}{width=\linewidth}
\begin{tabular}{|l|c|c|c|c|}
\hline
Algorithm & MAE & $<$ 20$^{\circ}$ & $<$ 25$^{\circ}$ & $<$ 30$^{\circ}$\\
\hline\hline
3DMM \cite{3DMM} & 26.3$^{\circ}$   & 4.3\% & 56.1\% & \textbf{89.4\%} \\
Pix2Vertex \cite{Sela_2017_ICCV} & 33.9$^{\circ}$ & 24.8\% & 36.1\% & 47.6\% \\
SfSNet\cite{SfSNet} & 25.5$^{\circ}$ & 43.6\% & 57.7\% & 68.7\% \\
SharinGAN (proposed)& \textbf{24.0$^{\circ}$} & \textbf{47.88}$\%$& \textbf{61.53}$\%$& 72.1$\%$\\
\hline
\end{tabular}
\end{adjustbox}
\caption{Quantitative results for Face Normal estimation on the test split of Photoface dataset \cite{Photoface}. All the listed methods are not fine-tuned on Photoface. The metrics MAE: Mean Angular Error and $< 20^\circ, 25^\circ, 30^\circ$ refer to the normals prediction accuracy for different thresholds.}
\vspace{-3mm}
\label{table:sn_results}
\end{table}

\textbf{Results} Table~\ref{table:sn_results} shows the quantitative performance of the estimated surface normals by our method on the test split of the Photoface dataset. 
With the proposed SharinGAN module, we were able to significantly improve over SfSNet on all the metrics.
In particular, we were able to significantly reduce the mean angular error metric by roughly 1.5$^{\circ}$.

\begin{figure}[!h]
    \begin{subfigure}{0.25\linewidth}{
        \includegraphics[width=\linewidth]{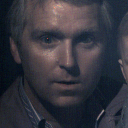}
        \includegraphics[width=\linewidth]{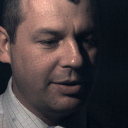}
        \includegraphics[width=\linewidth]{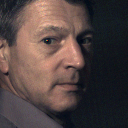}}
        \caption{Input Image}
    \end{subfigure}%
    ~
    \begin{subfigure}{0.25\linewidth}{
        \includegraphics[width=\linewidth]{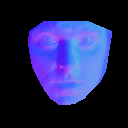}
        \includegraphics[width=\linewidth]{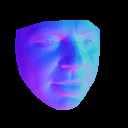}
        \includegraphics[width=\linewidth]{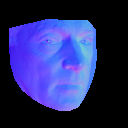}}
        \caption{GT}
    \end{subfigure}%
    ~
    \begin{subfigure}{0.25\linewidth}{
        \includegraphics[width=\linewidth]{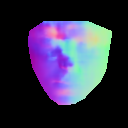}
        \includegraphics[width=\linewidth]{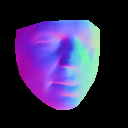}
        \includegraphics[width=\linewidth]{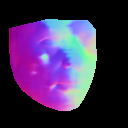}}
        \caption{SfSNet\cite{SfSNet}}
    \end{subfigure}%
    ~
    \begin{subfigure}{0.25\linewidth}{
        \includegraphics[width=\linewidth]{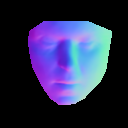}
        \includegraphics[width=\linewidth]{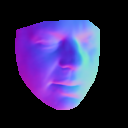}
        \includegraphics[width=\linewidth]{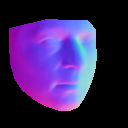}}
        \caption{SharinGAN}
    \end{subfigure}%
\caption{Qualitative comparisons of our method with SfSNet on the examples from the test set of Photoface dataset \cite{Photoface}. Our method generalizes much better to unseen data during training. }
\label{fig:FNE_SharinGAN_photoface}
\end{figure}
Additionally, Figure \ref{fig:FNE_SharinGAN_photoface}  depicts the qualitative comparison of our method with SfSNet on the test split of Photoface.
Both SfSNet and our pipeline are not finetuned on this dataset, and yet we were able to generalize better compared to SfSNet. 
This demonstrates the generalization capacity of the proposed SharinGAN to unseen data in training. 
%



\section{Ablation studies}

We carried out our ablation study using the KITTI \camready{and Make3D} dataset\camready{s} on monocular depth estimation.
We study the role of the SharinGAN module by removing it and training a primary network on the original synthetic and real data using  (\ref{eq:PT}).
We observe that the performance drops significantly as shown in Table \ref{tab:ablation_sharingan_training} \camready{and Table \ref{tab:ablation_sharingan_make3d}}. 
This shows the importance of the SharinGAN module that helps train the primary task network efficiently.

\camready{To demonstrate the role of reconstruction loss, we remove it and train our whole pipeline $\alpha_1 L_{adv} + \alpha_3 L_{T}$. We show the results on the testset of KITTI in the second row of Table~\ref{tab:ablation_sharingan_training} and on the testset of Make3D in the second row of Table~\ref{tab:ablation_sharingan_make3d}. For both the testsets, we can see the performance drop compared to our full model. Although the drop is smaller in the case of KITTI, it can be seen that the drop is significant for Make3D dataset that is unseen during training. This signifies the importance of reconstruction loss to generalize well to a domain not seen during training.}

%
%
%
\begin{table}[!htbp]
    \begin{adjustbox}{width=\linewidth}
    \centering
    \begin{tabular}{|c|c||c||c|c|c|}
    \hline
    \multicolumn{2}{|c||}{Components} & \multirow{2}{*}{Cap} & \multicolumn{3}{|c|}{Error Metrics, lower is better}\\
    \cline{1-2} \cline{4-6}
    SharinGAN & Reconstruction loss && Abs Rel & Sq Rel & RMSE \\
    \hline
        x & x & 70m & 0.476 & 8.058 & 9.449\\
        \checkmark & x & 70m & 0.401 & 5.318 & \textbf{8.377} \\
        \checkmark & \checkmark & 70m & \textbf{0.377} & \textbf{4.900}	& 8.388\\
    \hline
    \end{tabular}
    \end{adjustbox}
    \caption{Ablation study for monocular depth estimation to understand the role of the SharinGAN module and Reconstruction loss on the Make3D test dataset. We need both to get the best performance for this task.}
    \vspace{-5mm}
    \label{tab:ablation_sharingan_make3d}
\end{table}






\section{Conclusion}
Our primary motivation is to simplify the process of combining synthetic and real images in training.  Prior approaches often pick one domain and try to map images into it from the other domain.  Instead, we train a generator to map all images into a new, shared domain.  In doing this, we note that in the new domain, the images need not be indistinguishable to the human eye, only to the network that performs the primary task.  The primary network will learn to ignore extraneous, domain-specific information that is retained in the shared domain.

To achieve this, we propose a simple network architecture that rests on our new SharinGAN, which maps both real and synthetic images to a shared domain.  The resulting images retain domain-specific details that do not prevent the primary network from effectively combining training data from both domains.  We demonstrate this by achieving significant improvements over state-of-the-art approaches in two important applications, surface normal estimation for faces, and monocular depth estimation for outdoor scenes.  Finally, our ablation studies demonstrate the significance of the proposed SharinGAN in effectively combining synthetic and real data.




{\small
\bibliographystyle{ieee_fullname}
\bibliography{egbib}
}

\cleardoublepage

\section{More Implementation details}
The discriminator architecture we used for this work is:  $\{CBR(n,3,1),CBR(2*n,3,2)\}_{n=\{32,64,128,256\}}$, ${\{CBR(512,3,1),CBR(512,3,2)\}}_{K sets}$, $\{FcBR(1024)$, $FcBR(512)$, $Fc(1)\}$, where, CBR(out channels, kernel size, stride) = Conv + BatchNorm2d + ReLU and FcBR(out nodes) = Fully conncected + BatchNorm1D + ReLU and Fc is a fully connected layer. For face normal estimation, we do not use batchnorm layers in the discriminator. We use the value $K=2$ for MDE and $K=1$ for FNE. 


\textbf{Face Normal Estimation} We update the generator 3 times for each update of the discriminator, which in turn is  updated 5 times internally as per \cite{WGAN,WGANGP}. The generator learns from a new batch each time, while the discriminator trains on a single batch for 5 times.

\section{Experiments}
\textbf{Monocular Depth Estimation} We provide more qualitative results on the test set of the Make3D dataset \cite{make3D}. Figure \ref{fig:MDE_SharinGAN_make3d_supplementary} further demonstrates the generalization ability of our method compared to \cite{GASDA}.

\textbf{Face Normal Estimation} Figure \ref{fig:FNE_SharinGAN_syn_real} depicts the qualitative results on the CelebA \cite{celebA} and Synthetic \cite{SfSNet} datasets. The translated images corresponding to synthetic and real images look similar in contrast to the MDE task (Figure 4 of the paper). We suppose that for the task of MDE, regions such as edges are domain specific, and yet hold primary task related information such as depth cues, which is why SharinGAN modifies such regions. However, for the task of FNE, we additionally predict albedo, lighting, shading and a reconstructed image along with estimating normals. This means that the primary network needs a lot of shared information across domains for good generalization to real data. Thus the SharinGAN module seems to bring everything into a shared space, making the translated images $\{x_r^{sh}, x_s^{sh}\}$ look visually similar.

\begin{figure}[t]
    \centering
    \begin{subfigure}{0.25\linewidth}{
        \includegraphics[width=\linewidth]{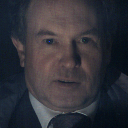}
        \includegraphics[width=\linewidth]{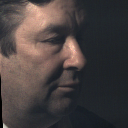}
        \includegraphics[width=\linewidth]{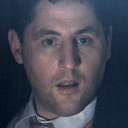}}
        \caption{Input Image}
    \end{subfigure}%
    ~
    \begin{subfigure}{0.25\linewidth}{
        \includegraphics[width=\linewidth]{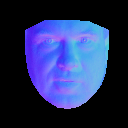}
        \includegraphics[width=\linewidth]{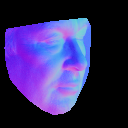}
        \includegraphics[width=\linewidth]{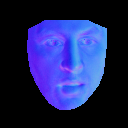}}
        \caption{GT}
    \end{subfigure}%
    ~
    \begin{subfigure}{0.25\linewidth}{
        \includegraphics[width=\linewidth]{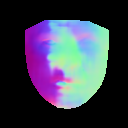}
        \includegraphics[width=\linewidth]{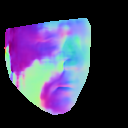}
        \includegraphics[width=\linewidth]{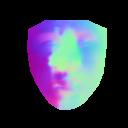}}
        \caption{SfSNet\cite{SfSNet}}
    \end{subfigure}%
    ~
    \begin{subfigure}{0.25\linewidth}{
        \includegraphics[width=\linewidth]{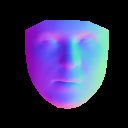}
        \includegraphics[width=\linewidth]{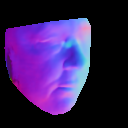}
        \includegraphics[width=\linewidth]{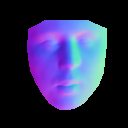}}
        \caption{SharinGAN}
    \end{subfigure}%
\caption{Additional Qualitative comparisons of our method with SfSNet on the examples from test set of the Photoface dataset \cite{Photoface}. Our method generalizes much better to unseen data during training. }
\label{fig:FNE_SharinGAN_photoface_supplementary}
\end{figure}

\begin{figure*}[h]
    \begin{subfigure}{0.25\linewidth}{
        \includegraphics[width=\linewidth]{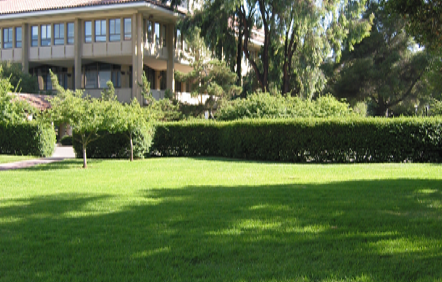}
        \includegraphics[width=\linewidth]{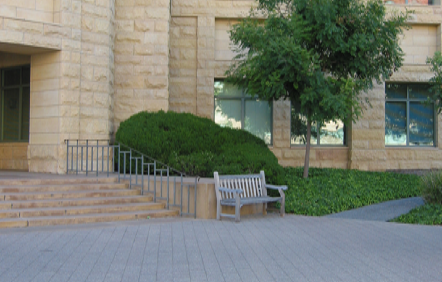}
        \includegraphics[width=\linewidth]{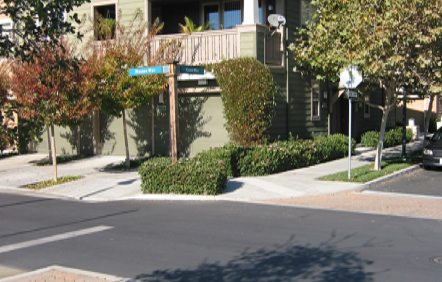}}
        \caption{Input Image}
    \end{subfigure}%
    ~
    \begin{subfigure}{0.25\linewidth}{
        \includegraphics[width=\linewidth]{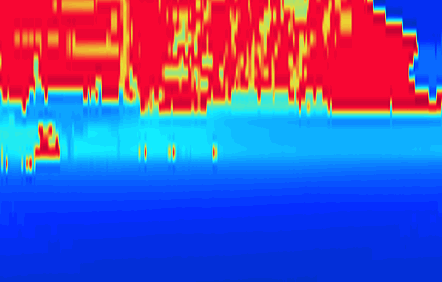}
        \includegraphics[width=\linewidth]{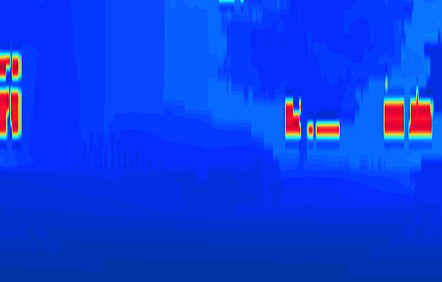}
        \includegraphics[width=\linewidth]{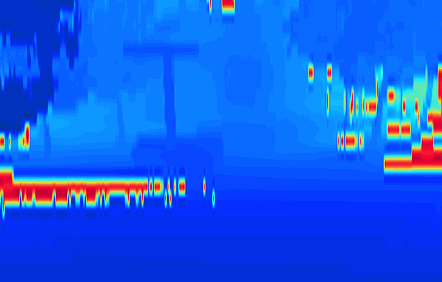}}
        \caption{Ground Truth}
    \end{subfigure}%
    ~
    \begin{subfigure}{0.25\linewidth}{
        \includegraphics[width=\linewidth]{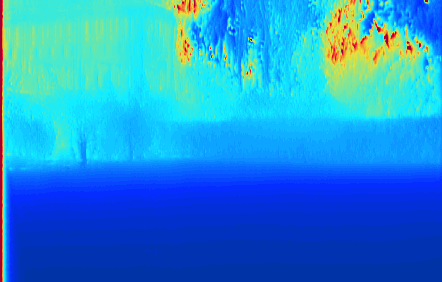}
        \includegraphics[width=\linewidth]{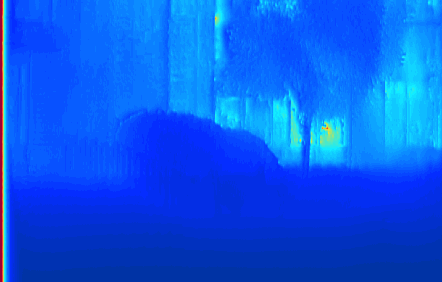}
        \includegraphics[width=\linewidth]{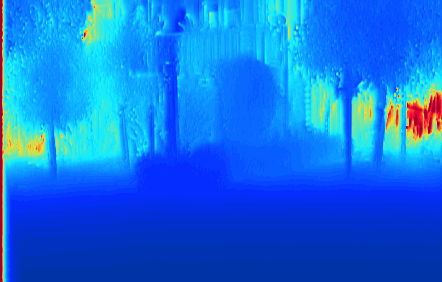}}
        \caption{GASDA\cite{GASDA}}
    \end{subfigure}%
    ~
    \begin{subfigure}{0.25\linewidth}{
        \includegraphics[width=\linewidth]{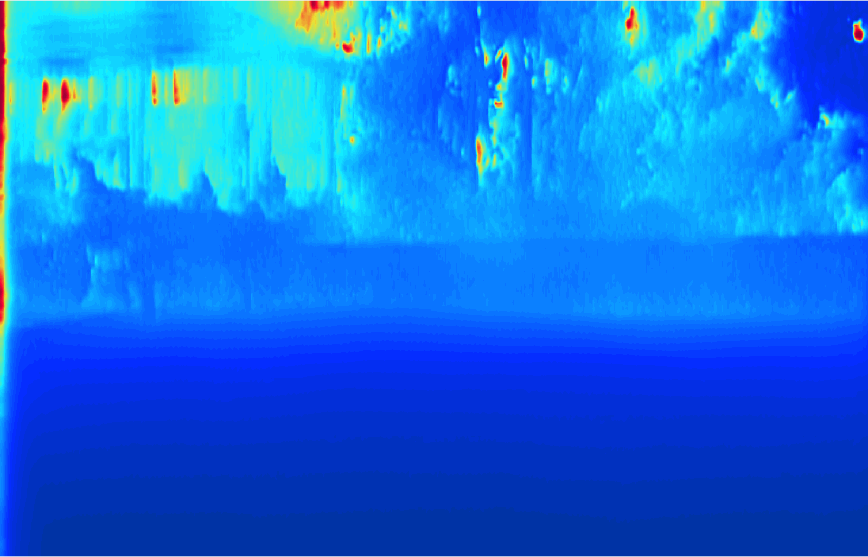}
        \includegraphics[width=\linewidth]{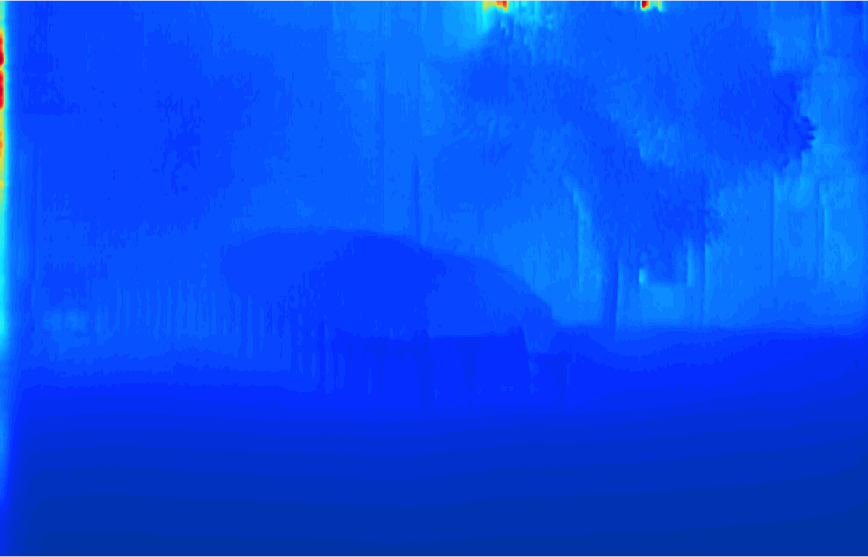}
        \includegraphics[width=\linewidth]{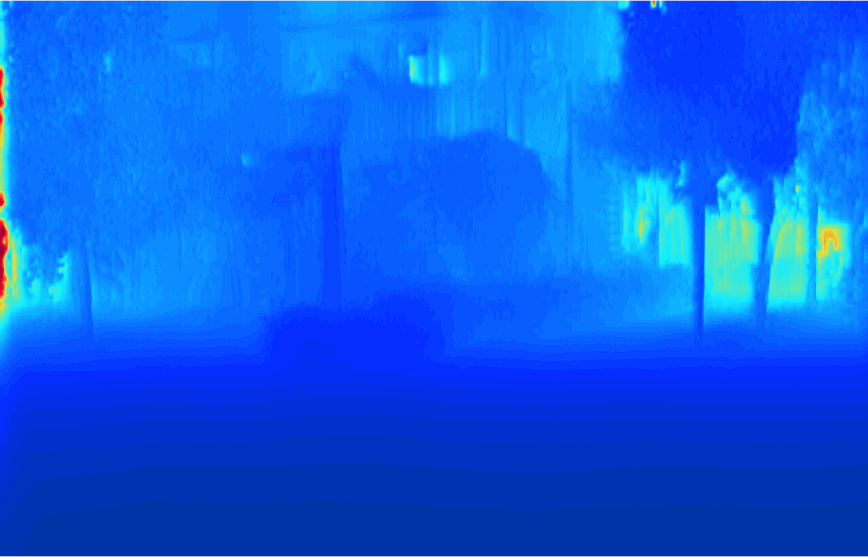}}
        \caption{SharinGAN}
    \end{subfigure}%
\caption{Additional Qualitative results on the test set of Make3D dataset \cite{make3D}. Our method is able to capture better depth estimates compared to \cite{GASDA} for all the examples.}
\label{fig:MDE_SharinGAN_make3d_supplementary}
\end{figure*}

\begin{figure*}
\centering
    \begin{subfigure}{\linewidth}
    \centering
        \begin{tabular}{cccccc}
            \includegraphics[width=0.15\linewidth]{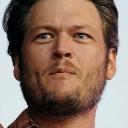}&
            \includegraphics[width=0.15\linewidth]{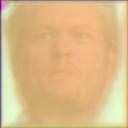}&
            \includegraphics[width=0.15\linewidth]{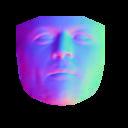}&
            \includegraphics[width=0.15\linewidth]{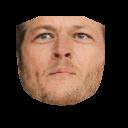}&
            \includegraphics[width=0.15\linewidth]{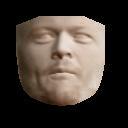}&
            \includegraphics[width=0.15\linewidth]{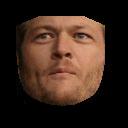}\\
            
            \includegraphics[width=0.15\linewidth]{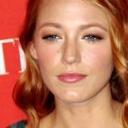}&
            \includegraphics[width=0.15\linewidth]{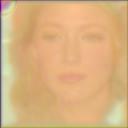}&
            \includegraphics[width=0.15\linewidth]{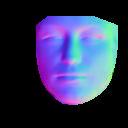}&
            \includegraphics[width=0.15\linewidth]{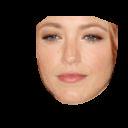}&
            \includegraphics[width=0.15\linewidth]{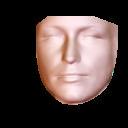}&
            \includegraphics[width=0.15\linewidth]{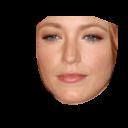}\\
            
            \includegraphics[width=0.15\linewidth]{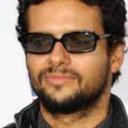}&
            \includegraphics[width=0.15\linewidth]{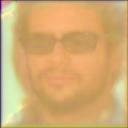}&
            \includegraphics[width=0.15\linewidth]{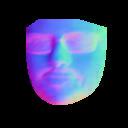}&
            \includegraphics[width=0.15\linewidth]{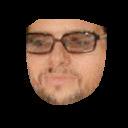}&
            \includegraphics[width=0.15\linewidth]{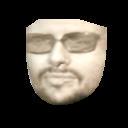}&
            \includegraphics[width=0.15\linewidth]{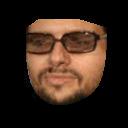}\\
            
            \includegraphics[width=0.15\linewidth]{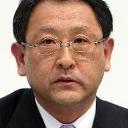}&
            \includegraphics[width=0.15\linewidth]{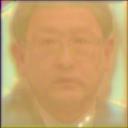}&
            \includegraphics[width=0.15\linewidth]{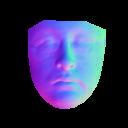}&
            \includegraphics[width=0.15\linewidth]{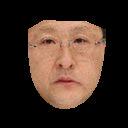}&
            \includegraphics[width=0.15\linewidth]{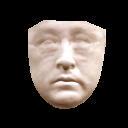}&
            \includegraphics[width=0.15\linewidth]{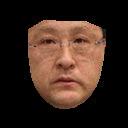}\\
            
            Input Image, $x_s$ & $x_s^{sh} = G(x_s)$ & Normal & Albedo & Shading & Reconstruction 
            \end{tabular}
            \caption{Qualitative results of our method on CelebA testset \cite{celebA}.}
        \end{subfigure}%
        

    \begin{subfigure}{\linewidth}
    \centering
        \begin{tabular}{cccccc}
            \includegraphics[width=0.15\linewidth]{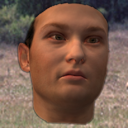}&
            \includegraphics[width=0.15\linewidth]{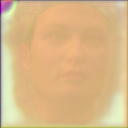}&
            \includegraphics[width=0.15\linewidth]{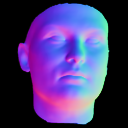}&
            \includegraphics[width=0.15\linewidth]{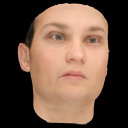}&
            \includegraphics[width=0.15\linewidth]{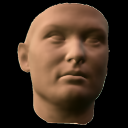}&
            \includegraphics[width=0.15\linewidth]{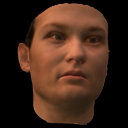}\\
            \includegraphics[width=0.15\linewidth]{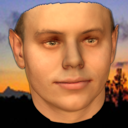}&
            \includegraphics[width=0.15\linewidth]{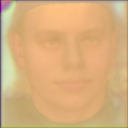}&
            \includegraphics[width=0.15\linewidth]{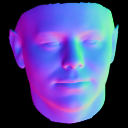}&
            \includegraphics[width=0.15\linewidth]{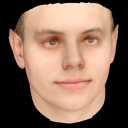}&
            \includegraphics[width=0.15\linewidth]{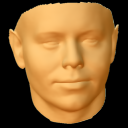}&
            \includegraphics[width=0.15\linewidth]{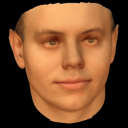}\\
            
            \includegraphics[width=0.15\linewidth]{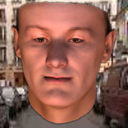}&
            \includegraphics[width=0.15\linewidth]{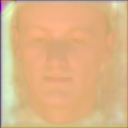}&
            \includegraphics[width=0.15\linewidth]{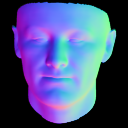}&
            \includegraphics[width=0.15\linewidth]{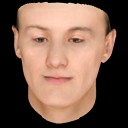}&
            \includegraphics[width=0.15\linewidth]{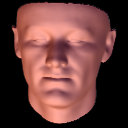}&
            \includegraphics[width=0.15\linewidth]{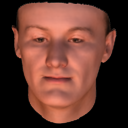}\\
            
            \includegraphics[width=0.15\linewidth]{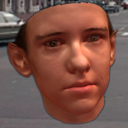}&
            \includegraphics[width=0.15\linewidth]{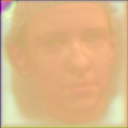}&
            \includegraphics[width=0.15\linewidth]{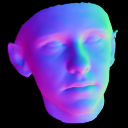}&
            \includegraphics[width=0.15\linewidth]{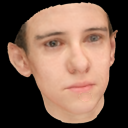}&
            \includegraphics[width=0.15\linewidth]{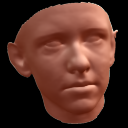}&
            \includegraphics[width=0.15\linewidth]{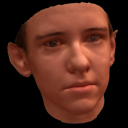}\\
            Input Image, $x_r$ & $x_r^{sh} = G(x_r)$ & Normal & Albedo & Shading & Reconstruction 
        \end{tabular}
        \caption{Qualitative results of our method on the synthetic data used in \cite{SfSNet}.}
        
    \end{subfigure}%
\caption{Qualitative results of our method on face normal estimation task. The translated images $x_r^{sh}, x_s^{sh}$ look reasonably similar for our task which additionally predicts albedo, lighting, shading and Reconstructed image along with the face normal.}
\label{fig:FNE_SharinGAN_syn_real}
\end{figure*}

Figure \ref{fig:FNE_SharinGAN_photoface_supplementary} depicts additional qualitative results of the predicted face normals for the test set of the Photoface dataset \cite{Photoface}. 

\begin{table}[!h]
    \centering
    \begin{tabular}{|c||c|c|c|}
    \hline
        Algorithm & top-1$\%$ & top-2$\%$ & top-3$\%$\\
        \hline
        SfSNet \cite{SfSNet} & 80.25 & 92.99 & 96.55\\
        SharinGAN & \textbf{81.83} & \textbf{93.88} & \textbf{96.69}\\
    \hline
    \end{tabular}
    \caption{Light classification accuracy on MultiPIE dataset \cite{Multipie}. Training with the proposed SharinGAN also improves lighting estimation along with face normals.}
        \vspace{-5mm}
    \label{tab:ablation_sharingan_multipie}
\end{table}
\textbf{Lighting Estimation} The primary network estimates not only face normals but also lighting.  We also evaluate this. Following a similar evaluation protocol as that of \cite{SfSNet}, Table \ref{tab:ablation_sharingan_multipie} summarizes the light classification accuracy on the MultiPIE dataset \cite{Multipie}. Since we do not have the exact cropped dataset that \cite{SfSNet} used, we used our own cropping and resizing on the original MultiPIE data: centercrop 300x300 and resize to 128x128. For a fair comparison, we used the same dataset to re-evaluate the lighting performance for \cite{SfSNet} and reported the results in Table \ref{tab:ablation_sharingan_multipie}. Our method not only outperforms \cite{SfSNet} on the face normal estimation, but also on lighting estimation.

\end{document}